\documentclass[journal]{IEEEtran}
\usepackage{graphicx}
\usepackage[utf8]{inputenc}
\usepackage{float}
\usepackage{booktabs}
\usepackage{indentfirst}
\usepackage{multirow}
\usepackage{enumitem}
\usepackage{amsmath}
\usepackage{amssymb}

\usepackage[normalem]{ulem} 
\usepackage{xcolor}

\usepackage{changes}
\setaddedmarkup{\textcolor{blue}{#1}}

\setdeletedmarkup{\textcolor{red}{\sout{#1}}}

\begin{document}

\title{\LARGE \bf
Shape Control of a Planar Hyper-Redundant Robot via Hybrid Kinematics-Informed and Learning-based Approach
}

\author{
Yuli Song,
        Wenbo Li,
        Wenci Xin,
        Zhiqiang Tang,
        Daniela Rus,
        and Cecilia Laschi
\thanks{This work funded in part by the National Research Foundation (NRF), Prime Minister’s Office, Singapore under its Campus for Research Excellence and Technological Enterprise (CREATE) programme. The Mens, Manus, and Machina (M3S) is an interdisciplinary research group (IRG) of the Singapore MIT Alliance for Research and Technology (SMART) centre, in part by Start-up Research Fund of Southeast University (RF1028625147), in part by National Key R\&D Program of China (2025YFE0214200) \textit{(Corresponding author: Wenci Xin (wenci.xin@smart.mit.edu))} }
\thanks{Y. Song, W. Li and W. Xin are with Singapore MIT Alliance for Research and Technology (SMART) centre, Singapore, and also with Department of Mechanical Engineering, National University of Singapore, Singapore}
\thanks{Z. Tang is with Jiangsu Key laboratory for design and Manufacturing of Precision Medicine equipment, School of Mechanical engineering, Southeast University, China}
\thanks{D. Rus is with CSAIL, Massachusetts Institute of Technology, Cambridge, USA, and also co-lead PI of Singapore MIT Alliance for Research and Technology (SMART) centre, Singapore}
\thanks{C. Laschi is with the Advanced Robotic Centre, National University of Singapore, Singapore; Department of Mechanical Engineering, National University of Singapore, Singapore, and also PI of Singapore MIT Alliance for Research and Technology (SMART) centre, Singapore}
} 

\maketitle

\begin{abstract}
Hyper-redundant robots offer high dexterity, making them good at operating in confined and unstructured environments. To extend the reachable workspace, we built a multi-segment flexible rack actuated planar robot. However, the compliance of the flexible mechanism introduces instability, rendering it sensitive to external and internal uncertainties. To address these limitations, we propose a hybrid kinematics-informed and learning-based shape control method, named SpatioCoupledNet. The neural network adopts a hierarchical design that explicitly captures bidirectional spatial coupling between segments while modeling local disturbance along the robot body. A confidence-gating mechanism integrates prior kinematic knowledge, allowing the controller to adaptively balance model-based and learned components for improved convergence and fidelity. The framework is validated on a five-segment planar hyper-redundant robot under three representative shape configurations. Experimental results demonstrate that the proposed method consistently outperforms both analytical and purely neural controllers. In complex scenarios, it reduces steady-state error by up to 75.5\% against the analytical model, and accelerates convergence by up to 20.5\% compared to the data-driven baseline. Furthermore, gating analysis reveals a state-dependent authority fusion, shifting toward data-driven predictions in unstable states, while relying on physical priors in the remaining cases. Finally, we demonstrate robust performance in a dynamic task where the robot maintains a fixed end-effector position while avoiding moving obstacles, achieving a precise tip-positioning accuracy with a mean error of 10.47~mm.
\end{abstract}

\section{INTRODUCTION}

Hyper-redundant robots, with their large number of controllable degrees of freedom (DoFs), are particularly well-suited for operation in confined environments and have become a key platform for studying advanced control strategies \cite{chirikjian1991hyper, mu2022hyper, niu2024shape}. Most existing hyper-redundant robots adopt either articulated motor–linkage architectures (e.g., snake robots \cite{yang2022snake}) or rear-mounted actuation schemes (e.g., tendon-driven hyper-redundant manipulators \cite{manara2025tendon}). However, their limited axial extension capability inherently constrains the reachable workspace and, in turn, restricts practical deployment. Inspired by the flexible rack actuation mechanism proposed in \cite{matsuda2022woodpecker}, we integrate multiple rack-actuated segments into a unified architecture, yielding a planar hyper-redundant robot with a substantially expanded workspace. This design, however, introduces a new challenge: coordinated whole-body control of multiple serially connected continuum segments.

\begin{figure}[!t]
    \centering
    \includegraphics[width=1.0\columnwidth]{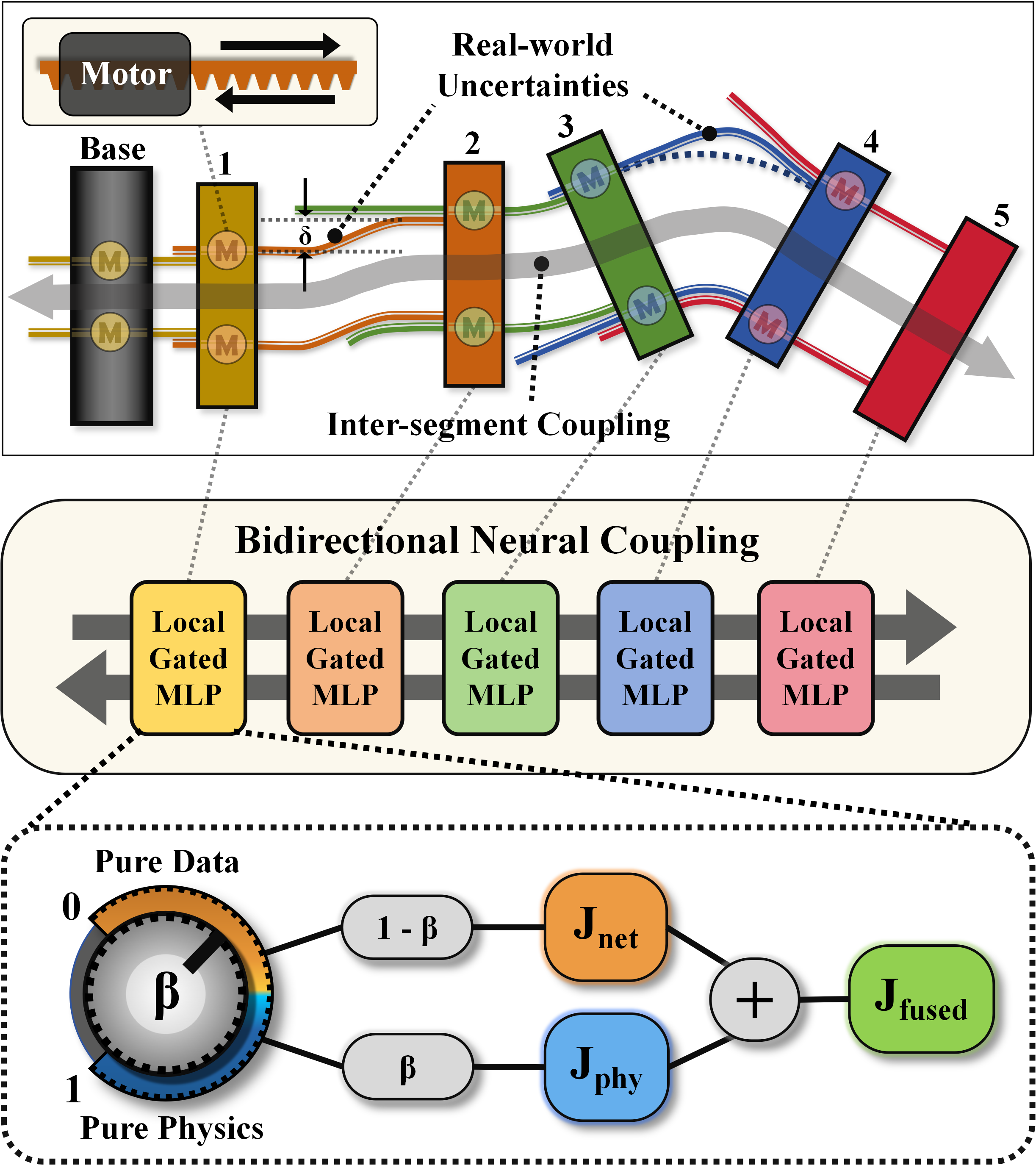} 
    \caption{\textbf{Overview of the proposed hybrid kinematics-informed and learning-based framework for shape control.} 
    (Top) Real-world physical uncertainties in a planar hyper-redundant robot. It illustrates how unmodeled physical effects (e.g., unexpected positioning shifts and non-ideal bending) and inter-segment force coupling complicate the actual motion. 
    (Middle) The SpatioCoupledNet architecture. By explicitly mirroring the physical mechanism, it employs a bidirectional neural structure to capture how forces propagate and interact across different segments. 
    (Bottom) The dynamic confidence gating mechanism. It acts as an adaptive dial ($\beta$) that smoothly balances control authority between the idealized physical model ($\mathbf{J}_{phy}$) and the data-driven neural prediction ($\mathbf{J}_{net}$). The resulting fused Jacobian ($\mathbf{J}_{fused}$) ensures high-fidelity closed-loop control, even under complex structural deformations.}
    \label{fig:teaser}
\end{figure}


A variety of analytical kinematic models have been developed for continuum robots. Simple geometric formulations, such as the PCC model \cite{webster2010design}, are computationally efficient, but their idealized assumptions limit their ability to capture the full range of deformations along the robot body \cite{zeng2021motion}. More advanced continuum mechanics approaches, including Cosserat rod theory \cite{alessi2024rod, tummers2023cosserat} and variable-curvature models \cite{huang2021kinematic}, provide higher modeling fidelity but are often computationally prohibitive for real-time control. For our architecture, however, strong inter-segment coupling effects make it inadequate to simply chain multiple kinematic models to describe the behavior of the hyper-redundant robot. In addition, the use of flexible racks increases the system’s sensitivity to external disturbances (e.g., friction) and internal uncertainties (e.g., material degradation), further exacerbating the discrepancy between model predictions and actual performance.


Learning-based approaches have gained significant attention in robotics for their ability to adapt to uncertainties and nonlinearities \cite{chen2024data}. However, purely data-driven controllers often lack physical grounding and can lead to over-actuation \cite{lutter2019deep}. To mitigate this, hybrid methods that perform residual learning on top of prior kinematic models have been proposed to reduce the mismatch between analytical predictions and real-world behavior \cite{gao2024sim, li2020model}. In our setting, however, the highly coupled interactions between segments are difficult to compensate for using residual learning layered on a simplified model. Even at the level of a single segment, the residual mapping exhibits high variance due to friction and material degradation, making it challenging to capture accurately.

In this work, we propose a novel kinematics-informed hybrid learning framework, named SpatioCoupledNet, for shape control of hyper-redundant robots subject to strong inter-segment coupling and high-variance disturbances. As illustrated in Fig. \ref{fig:teaser}, we adopt a hierarchical architecture. At the single-segment level, instead of directly performing residual learning to compensate for model errors, we introduce a confidence-gating mechanism that adaptively balances the reliance on the analytical kinematic model and the data-driven model. This design maximizes the utilization of prior knowledge while improving training efficiency and convergence stability. To enable real-time control, the computationally efficient PCC model is employed. Beyond the local gated module, a bidirectional recurrent structure is introduced across all segments to explicitly capture inter-segment coupling effects and long-range dependencies along the robot body.

The main contributions of this work are summarized as follows:

\begin{itemize}[wide=\parindent-0.95em, leftmargin=0pt, labelsep=0.5em, nosep]
    \item A kinematics-informed neural architecture, SpatioCoupledNet, that utilizes a bidirectional recurrent structure to explicitly model the spatial coupling and bidirectional propagation of internal stresses inherent in multi-segment hyper-redundant robots.
    \item A dimension-wise confidence gating strategy for state-dependent fusion, enabling the system to dynamically transition control authority between analytical kinematic priors and learned dynamics to compensate for unmodeled nonlinearities without sacrificing physical consistency.
    \item Comprehensive experimental validation on a custom hybrid robot demonstrating high-fidelity tracking in complex boundary configurations and real-time null-space obstacle avoidance. 
\end{itemize}

\section{ROBOTIC PLATFORM}
\begin{figure}[!t]
    \centering
    \includegraphics[width=1.0\columnwidth]{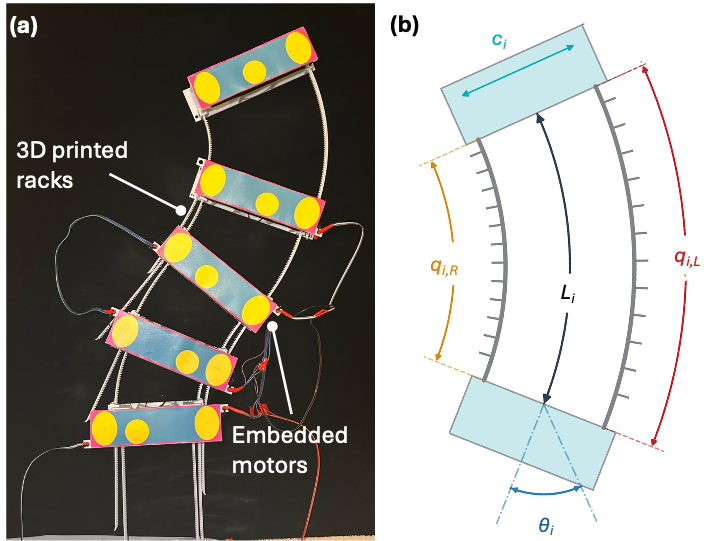} 
    \caption{\textbf{Kinematic parameterization of the proposed hyper-redundant robot.} (a) Planar hyper-redundant robotic platform used during the study with five continuum segments. (b) Local geometric model of the $i$-th segment based on the PCC assumption. The differential rack extensions $q_{i,L}$ and $q_{i,R}$ explicitly determine the bending angle $\theta_i$ and the central arc length $L_i$, where $c_i$ denotes the structural width.}
    \label{fig:kinematic_model}
\end{figure}

Figure \ref{fig:kinematic_model}(a) shows the robotic platform used during the study. Following the design in \cite{matsuda2022woodpecker}, each segment consists of two servo motors at its ends that control the extension of two 3D-printed PLA flexible racks. With the base fixed, the extension of two racks provides each segment with 2-DOF, forming a planar hyper-redundant system by stacking $N=5$ segments. Let $\mathbf{q} \in \mathbb{R}^{2N} \text{ [mm]}$ denote the joint actuation space, where $\mathbf{q}_i = [q_{i,L}, q_{i,R}]^T$ represents the left and right rack extensions for the $i$-th segment, and $\mathbf{X} \in \mathbb{R}^{3N}$ denotes the task-space poses. 

\subsection{Nominal Kinematic Modeling}

Under the PCC assumption \cite{webster2010design}, the local kinematics of the $i$-th segment explicitly map the joint-space extensions to the task-space pose $\mathbf{X}_i = [x_i, y_i, \theta_i]^T$. The bending angle $\theta_i$ and the central arc length $L_i$ are dictated by the differential and average extensions of the parallel racks:
\begin{equation}
    \theta_i = \frac{q_{i,L} - q_{i,R}}{c_i}, \quad L_i = \frac{q_{i,L} + q_{i,R}}{2}
\end{equation}
where $c_i$ denotes the structural width. By aggregating these local transformations along the serial backbone \cite{jones2006kinematics}, the instantaneous task-space displacement for the entire robot is then linearly approximated using the analytical physical Jacobian $\mathbf{J}_{phy}$:
\begin{equation}
    \Delta \mathbf{X}_{nom} \approx \mathbf{J}_{phy}(\mathbf{q}) \Delta \mathbf{q}
\end{equation}

To prevent hardware failure such as rack jamming or snapping due to excessive localized bending, the operational state space is strictly confined by highly non-linear inter-rack coupling constraints. The feasible actuation range for one rack is dynamically bounded by the current state of its partner:
\begin{equation}
    q_{i,L} \in \left[ f_{min}(q_{i,R}), f_{max}(q_{i,R}) \right]
\end{equation}
where $f_{min}$ and $f_{max}$ are empirically derived polynomial bounds. 

\begin{figure*}[t] 
    \centering
    \includegraphics[width=1.0\textwidth]{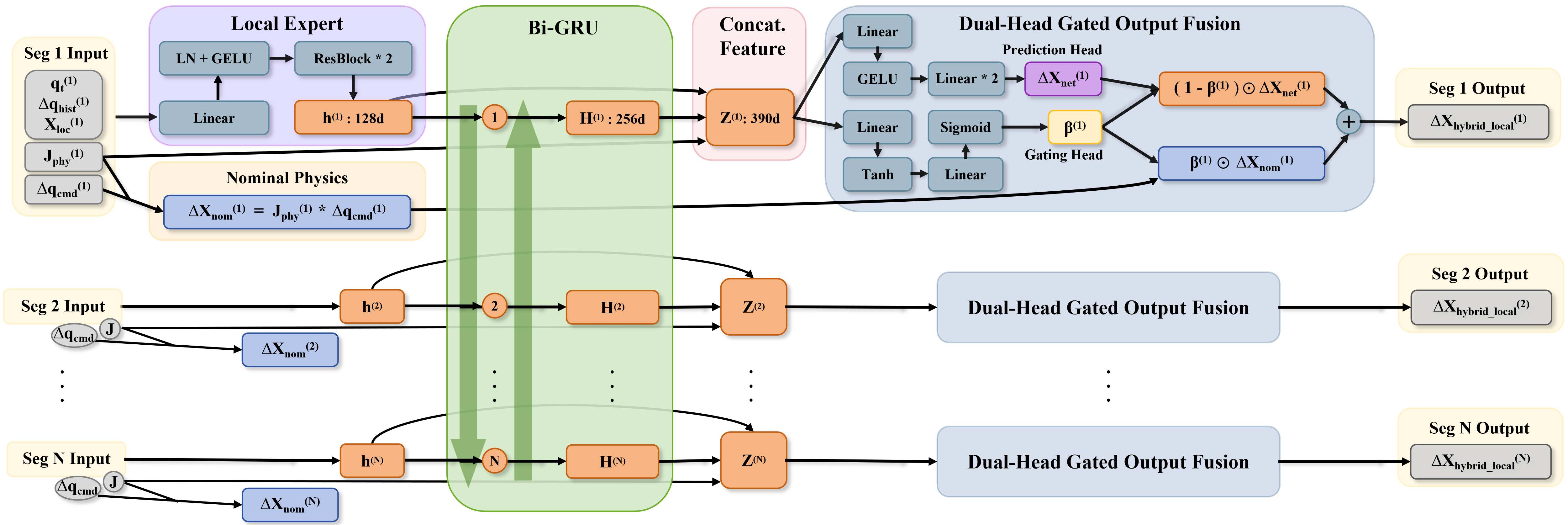}
    \caption{\textbf{Architectural hierarchy of SpatioCoupledNet for $N$-segment continuum robots.} The framework explicitly mirrors the robot's physical kinematics through three stages: 
    (a) Local Expert Stage where independent MLPs extract segment-specific features $\mathbf{h}^{(i)}$; 
    (b) Global Communication Stage utilizing a Bi-GRU to model bidirectional spatial coupling arising from distal actuation and proximal reaction forces; and 
    (c) Dual-Head Gated Fusion Stage where a dimension-wise confidence vector $\boldsymbol{\beta}^{(i)}$ adaptively balances analytical nominal physics and data-driven predictions. 
    }
    \label{fig:topo_impulse_net_hierarchy}
\end{figure*}

\subsection{Force Coupling and Unmodeled Effects}

The nominal PCC model operates under the idealized assumption that segmental deformations are purely geometric and physically isolated. In physical deployment, the robot behaves as a continuous elastic body subjected to distributed internal loads. When an actuation command is applied, the resulting localized bending moments and shear forces propagate bidirectionally along the continuous elastic backbone. Specifically, distal actuation induces proximal reactions, while proximal motions simultaneously alter the stress distribution in distal segments. Throughout this propagation, the effective force is further modulated by localized external friction, which varies dynamically with the robot's spatial configuration. Consequently, segments undergo force-induced passive deformations even when their corresponding actuators are rigidly locked.

Furthermore, as the segment bends, the unextended portion of a distal rack is forced to conform to the curvature of the proximal segments. This physical constraint induces structural interference: if the bending direction of a proximal segment obstructs the natural path of a distal rack, the rack undergoes forced bending before reaching its own segment. This interaction creates a complex mechanical coupling that propagates across any two segments. 

Therefore, the true mapping from joint space to task space is not a static geometric projection. Instead, it represents the solution to a highly coupled elastostatic boundary value problem. To formulate this mathematically, we lump the unmodeled structural elasticity, spatial force interference, and historical material hysteresis $\mathcal{H}$ into a state-dependent unmodeled dynamics term $\mathcal{F}_{unmodeled}$:
\begin{equation}
    \Delta \mathbf{X}_{real} = \mathbf{J}_{phy}(\mathbf{q}) \Delta \mathbf{q} + \mathcal{F}_{unmodeled}(\mathbf{q}, \Delta \mathbf{q}, \mathcal{H})
\end{equation}

The fundamental control objective is to compute an optimal actuation increment $\Delta \mathbf{q}^*$ that minimizes the task-space tracking error $\| \mathbf{X}_{tgt} - \mathbf{X}_{real} \|$ subject to the non-linear operational boundaries $q_{i,L} \in [ f_{min}(q_{i,R}), f_{max}(q_{i,R}) ]$. The core challenge is that these unmodeled dynamics $\mathcal{F}_{unmodeled}$ are analytically intractable and strongly correlated across adjacent segments. Attempting to directly isolate and learn this volatile discrepancy—as in traditional residual compensation methodologies—often leads to erratic control signals. Particularly in near-neutral configurations lacking sustained structural tension, or in over-bent states where localized rack instability, the mechanism exhibits unpredictable parasitic translations and non-uniform curvatures. This highly coupled physical reality necessitates a kinematics-aware framework to safely holistically compensate for these underlying mechanics through dynamic authority fusion.

\section{HYBRID KINEMATICS-INFORMED AND LEARNING-BASED FRAMEWORK}

To achieve high-precision shape control under the aforementioned complex force couplings, we propose a kinematics-informed and
learning-based framework. At its core, the system relies on a spatial-coupling neural architecture and a dynamic confidence gating mechanism. To ensure long-term robustness against hardware degradation and environmental shifts, this controller is supported by a boundary-aware streamlined data curation pipeline. 

\subsection{SpatioCoupledNet: Kinematics-Informed Architecture}

Unlike rigid mechanisms, unmodeled dynamics $\mathcal{F}_{unmodeled}$ in coupled mechanisms propagate spatially. To capture this, SpatioCoupledNet explicitly mirrors the $N$-segment physical kinematics via a local-to-global hierarchy. 

At step $t$, the $i$-th segment's state is parameterized as a 15-dimensional vector:
\begin{equation}
    \mathbf{s}_t^{(i)} = \left[ \mathbf{q}_t^{(i)}, \Delta\mathbf{q}_{hist}^{(i)}, \mathbf{X}_{loc}^{(i)}, \Delta\mathbf{q}_{cmd}^{(i)}, \text{vec}(\mathbf{J}_{phy}^{(i)}) \right]^T
\end{equation}
These features capture distinct physical phenomena: $\mathbf{q}_t^{(i)}$ and $\mathbf{X}_{loc}^{(i)}$ anchor configuration-dependent kinematics; $\Delta\mathbf{q}_{hist}^{(i)}$ provides historical context for non-linear friction and hysteresis; the flattened $\mathbf{J}_{phy}^{(i)}$ supplies an explicit analytical prior; and $\Delta\mathbf{q}_{cmd}^{(i)}$ is the proposed actuation step.

The network processes these features through three integrated stages that reflect the physical interactions of the robot. Initially, rather than relying on a centralized mapping, each segment utilizes an independent expert module for local feature extraction. Sequentially stacked linear layers, layer normalization, GELU activations, and residual blocks extract segment-specific representations $\mathbf{h}^{(i)} \in \mathbb{R}^{128}$ while effectively mitigating vanishing gradients.

To capture the global inter-segment dependencies, these local representations $\mathbf{h}^{(i)}$ are subsequently fed into a Bidirectional Gated Recurrent Unit (Bi-GRU) \cite{schuster1997bidirectional} sequence model. This bidirectional recurrent structure is designed to mirror the complex force coupling inherent in the multi-segment robot. The Bi-GRU effectively captures these concurrent interactions: it models how proximal bending obstructs the natural path of distal racks, inducing forced bending, while simultaneously accounting for how actuation loads and external friction propagate through the continuous backbone. By processing the segment sequence from both directions, the network extracts a latent representation $\mathbf{H}^{(i)}$ that is fully conditioned on the global mechanical state of the entire mechanism.

Finally, a dimension-wise hybrid gating mechanism bridges the analytical and neural domains. The concatenated feature $\mathbf{z}^{(i)} = [\mathbf{h}^{(i)}, \mathbf{H}^{(i)}, \text{vec}(\mathbf{J}_{phy}^{(i)})] \in \mathbb{R}^{390}$ feeds two parallel MLP heads: a prediction head regressing the data-driven full displacement $\Delta \mathbf{X}_{net}^{(i)}$, and a confidence gate outputting a trust vector $\boldsymbol{\beta}^{(i)} \in [0, 1]^3$. The final displacement is fused via a dimension-wise convex combination:
\begin{equation}
    \Delta \mathbf{X}_{hybrid}^{(i)} = \boldsymbol{\beta}^{(i)} \odot \Delta \mathbf{X}_{nom}^{(i)} + (\mathbf{1} - \boldsymbol{\beta}^{(i)}) \odot \Delta \mathbf{X}_{net}^{(i)}
\end{equation}
where $\odot$ is the Hadamard product. This soft-gated system independently assigns confidence to translation and rotation. The gate is initialized with a bias toward the nominal physics $\Delta \mathbf{X}_{nom}$ to ensure operational safety and provide an analytical prior that accelerates convergence. During execution, predictive authority dynamically shifts to the neural prediction $\Delta \mathbf{X}_{net}$ to compensate for complex unmodeled dynamics, such as external friction.

\subsection{Physics-Constrained Loss and Differentiable Kinematics}

Given the coupled nature of the robot, kinematic error propagation causes minor angular deviations at proximal segments to amplify into significant task-space deviations at the distal end. To penalize this compounding effect, the loss formulation integrates a Differentiable Forward Kinematics (FK) layer, ensuring the network respects both local joint-space dynamics and global task-space consistency.

\begin{figure}[!b]
    \centering
    \includegraphics[width=0.48\textwidth]{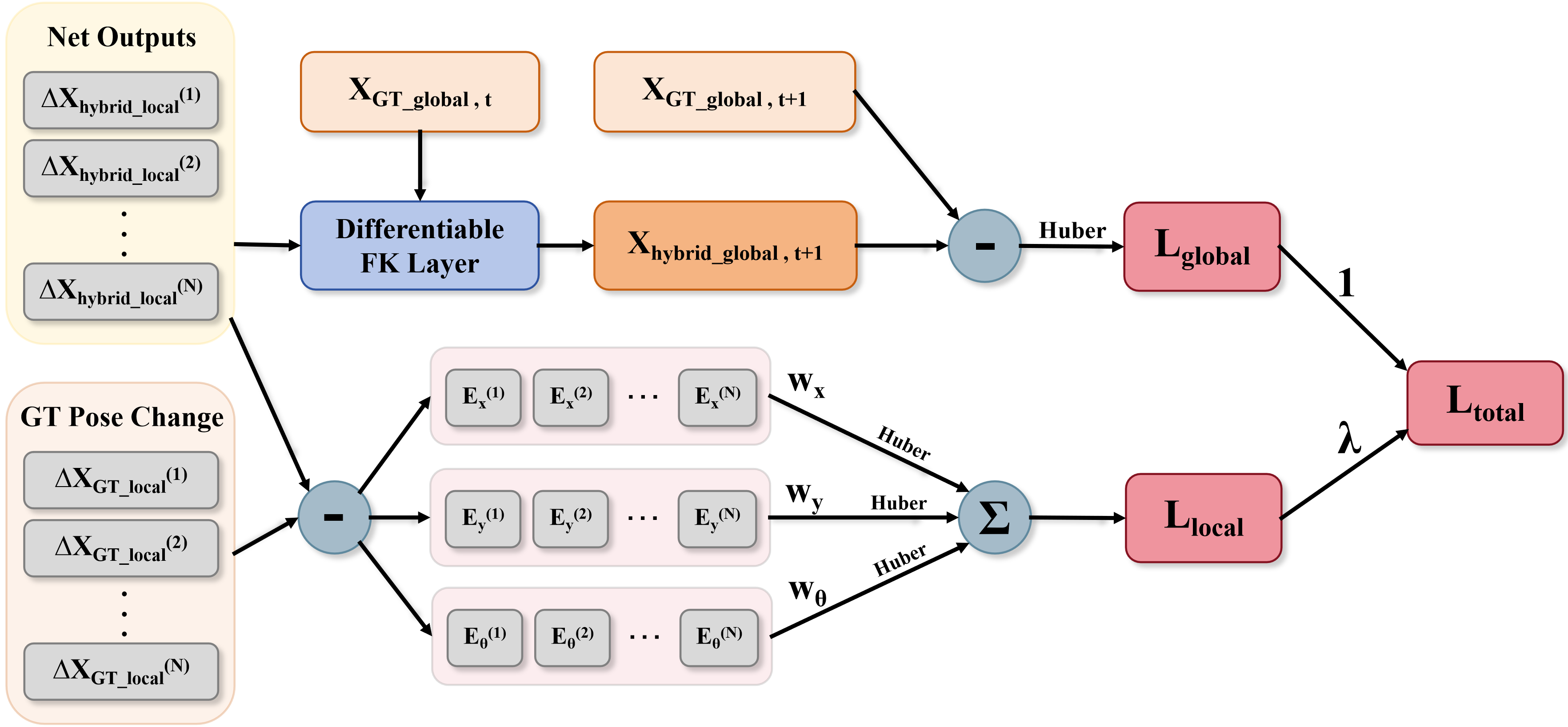}
    \caption{\textbf{Schematic of the physics-constrained multi-objective loss function.} The total loss $\mathcal{L}_{total}$ is composed of: (1) a local coordinate loss $\mathcal{L}_{local}$ penalizing deviations in segment-wise incremental poses; and (2) a global shape loss $\mathcal{L}_{global}$ enforced through a Differentiable Forward Kinematics (FK) layer. Both objectives utilize a Huber penalty to ensure robustness against perception outliers.}
    \label{fig:loss_structure}
\end{figure}

For the segment-level supervision, let $\mathbf{e}^{(i)} = \Delta \mathbf{X}_{hybrid\_local}^{(i)} - \Delta \mathbf{X}_{GT\_local}^{(i)}$ denote the local residual error vector for the $i$-th segment. The local coordinate loss $\mathcal{L}_{local}$ computes the weighted sum of these Huber penalties, heavily skewed towards orientation accuracy:
\begin{equation}
    \mathcal{L}_{local} = \sum_{i=1}^N \left( w_x L_{\delta}(e_x^{(i)}) + w_y L_{\delta}(e_y^{(i)}) + w_{\theta} L_{\delta}(e_{\theta}^{(i)}) \right)
\end{equation}
Based on the robot's physical characteristics, we enforce $w_{\theta} \gg \{w_x, w_y\}$ to strictly penalize angular divergence. 

Subsequently, the local hybrid predictions are integrated through the Differentiable FK layer. As illustrated in Fig. \ref{fig:loss_structure}, this layer takes the previous ground-truth global pose $\mathbf{X}_{GT\_global, t}$ as the kinematic base to compute the predicted global task-space poses $\mathbf{X}_{hybrid\_global, t+1}$. The global shape loss $\mathcal{L}_{global}$ then applies the same Huber penalty to the deviation between the predicted and ground-truth global states, forcing the network to respect global physical plausibility:
\begin{equation}
    \mathcal{L}_{global} = L_{\delta}(\mathbf{X}_{hybrid\_global, t+1} - \mathbf{X}_{GT\_global, t+1})
\end{equation}
Consistent with the fusion weights illustrated in the computational graph, the total loss is defined as:
\begin{equation}
    \mathcal{L}_{total} = \mathcal{L}_{global} + \lambda \mathcal{L}_{local}
\end{equation}

\begin{figure*}[htbp]
    \centering
    \includegraphics[width=1.0\textwidth]{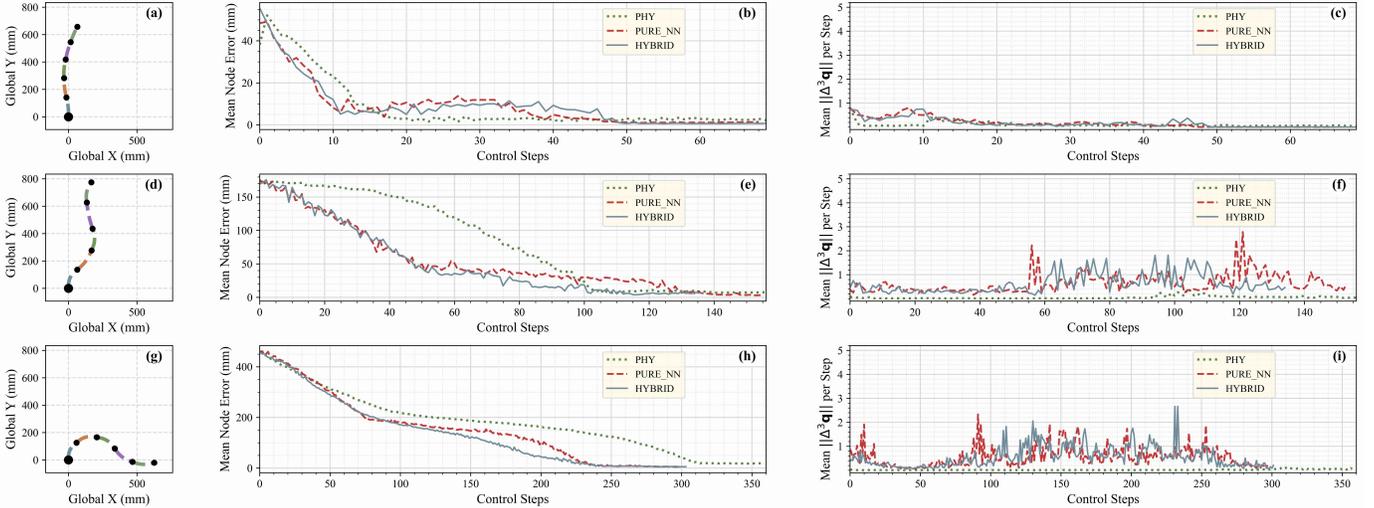} 
    \caption{\textbf{Tracking performance across three configuration difficulties in the tracking experiment.} The three rows correspond to the Easy (a--c), Medium (d--f), and Extreme (g--i) target configurations, respectively. \textbf{Left column (a, d, g):} The target spatial configurations of the robot. \textbf{Middle column (b, e, h):} The evolution of the mean node error ($e_{mean}$) over control steps, illustrating the convergence speed and steady-state accuracy. \textbf{Right column (c, f, i):} Control action chattering, quantified by the third-order difference of joint positions ($\|\Delta^3 \mathbf{q}\|$) per step.
    }
    \label{fig:tracking_results}
\end{figure*}

\subsection{Confidence Gating and Closed-Loop Deployment}

During closed-loop deployment, we extract a local Neural Jacobian $\mathbf{J}_{net} = \frac{\partial (\Delta \mathbf{X}_{net})}{\partial (\Delta \mathbf{q}_{cmd})}$ via real-time forward numerical perturbation of the network inputs. 

To avoid over-constraining the actuation space, the Damped Least Squares (DLS) controller \cite{nakamura1986inverse} operates exclusively within the 2D translational subspace. Although the orientation $\theta$ is omitted in the inverse solver, its accurate estimation during training remains necessary because proximal angular errors compound into distal translational deviations. Let $\mathbf{P} \in \mathbb{R}^{2N}$ denote the positional components $(X, Y)$ extracted from the full task-space state $\mathbf{X}$. Consequently, we extract the corresponding translational sub-matrices $\mathbf{J}_{phy, P}$ and $\mathbf{J}_{net, P} \in \mathbb{R}^{2N \times 2N}$ from their full-state counterparts. The fused spatial Jacobian $\mathbf{J}_{fused}$ is defined as:
\begin{equation}
    \mathbf{J}_{fused} = \mathbf{B}_{\beta} \mathbf{J}_{phy, P} + (\mathbf{I} - \mathbf{B}_{\beta}) \mathbf{J}_{net, P}
\end{equation}
where $\mathbf{B}_{\beta} \in \mathbb{R}^{2N \times 2N}$ is a diagonal matrix constructed from the translational dimensions of the confidence vectors $\boldsymbol{\beta}^{(i)}$. 

This gating mechanism dynamically reweights control authority based on the robot's spatial and kinematic state. At proximal segments or in smooth configurations, $\mathbf{B}_{\beta}$ prioritizes $\mathbf{J}_{phy, P}$ to leverage analytical stability and ensure directionally consistent gradients. Conversely, for distal segments or near-neutral states where unmodeled friction and hysteresis dominate, the controller shifts reliance toward $\mathbf{J}_{net, P}$. By learning through the Bi-GRU, this neural component captures the complex inter-segment couplings, such as the forced bending of distal racks and the concurrent propagation of internal stresses that analytical priors typically omit. Throughout the workspace, the physical prior serves as a structural regularizer, ensuring physically-plausible joint velocities.

To counteract asynchronous visual latency $\Delta t_{delay}$ during execution, a first-order Taylor expansion utilizes the fused Jacobian and instantaneous joint velocities $\dot{\mathbf{q}}$ to predict the current unobservable positional state:
\begin{equation}
    \mathbf{P}_{current} = \mathbf{P}_{vision} + \mathbf{J}_{fused} \dot{\mathbf{q}} \Delta t_{delay}
\end{equation}

Finally, target joint commands are computed using the adaptive DLS inverse solve:
\begin{equation}
    \Delta \mathbf{q}^* = (\mathbf{J}_{fused}^T \mathbf{W} \mathbf{J}_{fused} + \lambda_{dls} \mathbf{I})^{-1} \mathbf{J}_{fused}^T \mathbf{W} \mathbf{e}_{step}
\end{equation}
where $\mathbf{e}_{step}$ is the instantaneous Cartesian tracking error. To maximize tracking efficiency, $\mathbf{W}$ is formulated as a dynamic Gaussian weight matrix. The peak of this Gaussian distribution autonomously tracks the segment index with the highest positional error, effectively redistributing the actuation effort across the structural backbone. The resulting commands are clamped by bounding equations to strictly prevent mechanical self-collision and excessive bending.


\section{EXPERIMENTS AND RESULTS}

All experiments were conducted on the custom 5-segment continuum robot driven by servos. The control framework was implemented on a workstation with an Intel Core i7-12700H CPU. For real-time feedback, an overhead camera provides state estimation at a frame rate of 30~Hz, while the control loop operates at 50~Hz.

\subsection{Tracking Performance}

To evaluate the proposed framework, physical tracking experiments were conducted. The robot was initialized from a predefined state with all actuation
components set to q = 70 and tracked targets across three difficulty levels: Easy (mild bending), Medium (moderate coupling), and Extreme (severe structural deformation). We benchmark our hybrid kinematics-informed controller (HYBRID) against a purely analytical model (PHY, where $\beta = 1$) and a purely data-driven neural network (PURE\_NN, where $\beta = 0$). Tracking performance is analyzed using four metrics: steady-state mean error ($e_{mean}$), convergence speed ($T_{95\%}$, defined as the control steps required to reach 95\% steady-state convergence), action chattering (mean control jerk, $\|\Delta^3 \mathbf{q}\|$ per step), and total actuation cost (cumulative joint displacement, $\sum \|\Delta \mathbf{q}\|$).

\begin{table}[htbp]
    \centering
    \caption{Quantitative Tracking Performance Across Configurations. Bold values indicate the best (lowest) performance.}
    \label{tab:tracking_metrics}
    \setlength{\tabcolsep}{7pt} 
    \begin{tabular}{@{}llcccc@{}}
    \toprule
    \textbf{Config.} & \textbf{controller} & \boldmath{$e_{\text{mean}}$} & \boldmath{$T_{95\%}$} & \textbf{Chatter.} & \textbf{Cost} \\ 
    & & (mm) $\downarrow$ & (steps) $\downarrow$ & (mm) $\downarrow$ & (mm) $\downarrow$ \\ \midrule
    
    \multirow{3}{*}{\textbf{Easy}} 
    & PHY      & 2.13  & \textbf{23}  & \textbf{0.095} & 408.9 \\
    & PURE\_NN & 1.18  & 47  & 0.179 & 378.5 \\
    & HYBRID   & \textbf{0.74}  & 47  & 0.154 & \textbf{371.6} \\ \midrule
    
    \multirow{3}{*}{\textbf{Medium}} 
    & PHY      & 7.53  & 106 & \textbf{0.062} & 2078.0 \\
    & PURE\_NN & 6.32  & 127 & 0.646 & 1749.0 \\
    & HYBRID   & \textbf{5.84}  & \textbf{101} & 0.622 & \textbf{1558.6} \\ \midrule
    
    \multirow{3}{*}{\textbf{Extreme}} 
    & PHY      & 26.76 & 311 & \textbf{0.017} & \textbf{3285.1} \\
    & PURE\_NN & 7.12  & 230 & 0.564 & 3831.1 \\
    & HYBRID   & \textbf{6.55}  & \textbf{222} & 0.557 & 3603.8 \\ \bottomrule
    \end{tabular}
\end{table}

The quantitative results across different configurations are summarized in Table~\ref{tab:tracking_metrics}, with the corresponding dynamic responses illustrated in Fig.~\ref{fig:tracking_results}. In Easy configurations, the PCC assumption holds well. The PHY controller achieves rapid convergence ($T_{95\%} = 23$) with high smoothness (0.095~mm), but leaves a 2.13~mm steady-state error. As the target approaches workspace boundaries (Medium \& Extreme), structural elasticity and inter-segment coupling exacerbate the model-plant mismatch. While the PHY model remains capable of driving the system toward the target, its convergence slows down significantly (e.g., $T_{95\%}$ increases to 311 in the Extreme case). Furthermore, unmodeled spatial coupling causes the analytical Jacobian to deviate from actual kinematics, leading to a noticeable increase in steady-state error ($e_{mean} = 26.76$~mm in Extreme). This indicates the inherent difficulty of using purely analytical models for high-fidelity control in complex boundary states.

Learning-based models effectively compensate for these unmodeled effects, drastically reducing Cartesian errors in extreme states. However, this active compensation inherently requires high-frequency localized control adjustments, resulting in a comparable increase in control chattering for both PURE\_NN and HYBRID (Chatter. $>$ 0.5~mm in both Medium and Extreme configurations). The critical distinction lies in convergence efficiency and actuation effort. Without physical grounding, the PURE\_NN approach generates less directed exploration gradients in unstructured state spaces, leading to slower convergence ($T_{95\%} = 230$ in Extreme) and the highest overall actuation effort (Cost = 3831.1~mm in Extreme). Conversely, our HYBRID controller anchors neural predictions with a kinematics-informed physical prior. This integration acts as a crucial regularizer for the learning process, effectively accelerating convergence ($T_{95\%}$ reduced to 101 in Medium and 222 in Extreme) and decreasing the actuation cost (1558.6~mm in Medium and 3603.8~mm in Extreme), while maintaining the lowest steady-state error ($e_{mean} = 5.84$~mm in Medium and 6.55~mm in Extreme). This objectively validates that the physical prior accelerates convergence and balances high-fidelity tracking with operational efficiency.

\subsection{Confidence Gating Analysis}

\begin{figure}[htbp]
    \centering
    \includegraphics[width=1.0\columnwidth]{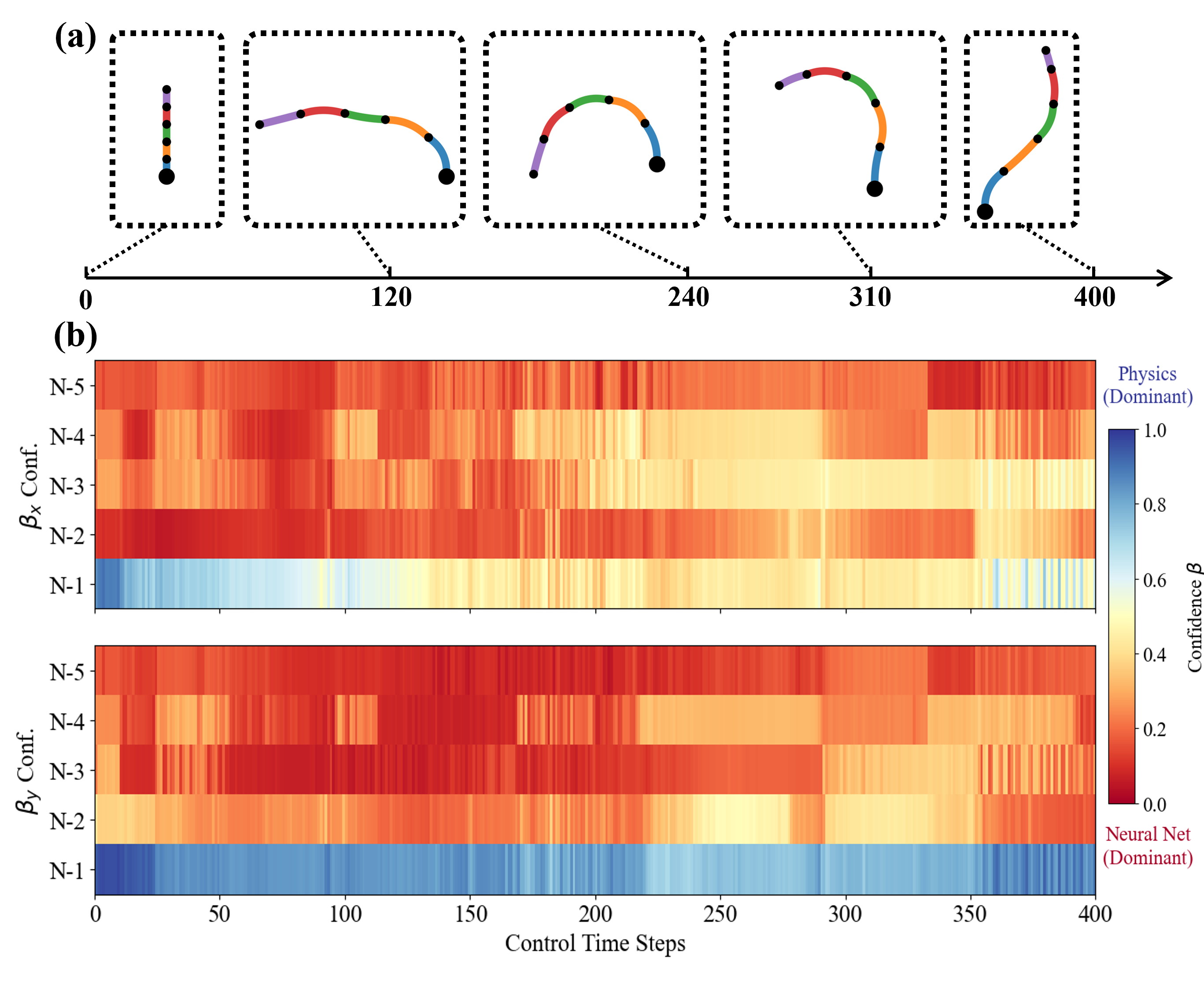} 
    \caption{\textbf{Evolution of Robot Configuration and Confidence Gating.} (a) Snapshots of spatial configurations at key intervals ($T=\{0, 120, 240, 310, 400\}$) during a continuous trajectory. (b) Real-time evolution of the dimension-wise confidence gate $\boldsymbol{\beta}$ for all five segments ($N=1$ to $N=5$) over 400 control steps. The heatmaps illustrate the dynamic shift of control authority between the analytical PCC prior ($\boldsymbol{\beta} \to 1$, blue) and the data-driven neural prediction ($\boldsymbol{\beta} \to 0$, red).}
    \label{fig:gating_evolution}
\end{figure}

To understand the underlying mechanism of the hybrid controller, we analyzed the real-time evolution of the dimension-wise confidence gate $\boldsymbol{\beta}$ during a continuous, multi-waypoint trajectory tracking task. For this specific evaluation, the robot was uniformly initialized with all actuation components set to $q = 10$. The target trajectory was designed to drive the mechanism through a diverse sequence of morphological phases. It initially guided the robot to extend leftward (e.g., around T=120); subsequently commanded a severe downward retroflexion, forcing the end-effector to curl back beyond the y-coordinate of the robot's base (e.g., around T=240); and finally prompted the mechanism to progressively unfurl into a smooth, extended posture with decreasing curvature (e.g., approaching T=400). Fig. \ref{fig:gating_evolution} visualizes the robot's spatial configuration (a) alongside the corresponding $\beta_x$ and $\beta_y$ heatmaps (b) for all five segments over the control steps. Recall that $\beta \to 1$ assigns predictive authority to the analytical PCC prior, while $\beta \to 0$ shifts dominance to the neural network prediction.

A distinct spatial hierarchy is immediately evident from the heatmaps. For the proximal base segment (Node 1), $\beta_y$ remains consistently high (near 1.0) throughout the entire trajectory, while $\beta_x$ initializes high before stabilizing around 0.5. This indicates that the base segment, which is rigidly grounded and suffers minimally from accumulated proximal reaction forces, adheres closely to nominal physical laws, particularly in its longitudinal extension. 

In stark contrast, the distal segments (Nodes 2 through 5) exhibit predominantly low confidence values ($\beta < 0.5$, denoted by warmer colors) for the majority of the task. As stress propagates along the coupled backbone, distal nodes are subjected to mechanical hysteresis, external friction, and compounded spatial coupling. The theoretical PCC model inherently fails to capture these highly non-linear, state-dependent elastic distortions. Consequently, the gating mechanism autonomously recognizes this analytical degradation and dynamically allocates predictive authority to the neural network, successfully compensating for the complex unmodeled effects.

Furthermore, the gating values demonstrate a strong temporal correlation with the instantaneous structural configuration. For instance, around T=120, the distal segments exhibit small, alternating curvatures. In such near-neutral states, the lack of sustained structural tension amplifies the effects of external friction, causing the physical mechanism to suffer from parasitic translations and non-uniform curvature. Consequently, the geometry deviates significantly from idealized predictions, as reflected by the universally low $\boldsymbol{\beta}$ values. Conversely, as the robot transitions into a smooth configuration (e.g., spanning from T=240 to T=400), it adopts a naturally curved posture, maintaining internal structural tension. This active tensioning physically constrains the flexible components into arcs that closely adhere to the PCC assumption, resulting in a widespread increase in $\boldsymbol{\beta}$ values. This transient shift back to the analytical prior objectively validates that SpatioCoupledNet does not blindly override the underlying physics; rather, it performs a targeted, state-dependent authority fusion that maximizes both tracking fidelity and structural safety.

\subsection{Real-time Obstacle Avoidance Demonstration}

To further validate the practical utility and dexterity of the proposed HYBRID controller, we designed a real-time obstacle avoidance task leveraging the hyper-redundant nature of the robot. The inherent kinematic redundancy allows the robot to execute null-space motions \cite{Alain1977NullSpace} by reconfiguring its internal structural shape to evade collisions without altering the target end-effector pose.

\begin{figure}[!t]
    \centering
    \includegraphics[width=1.0\columnwidth]{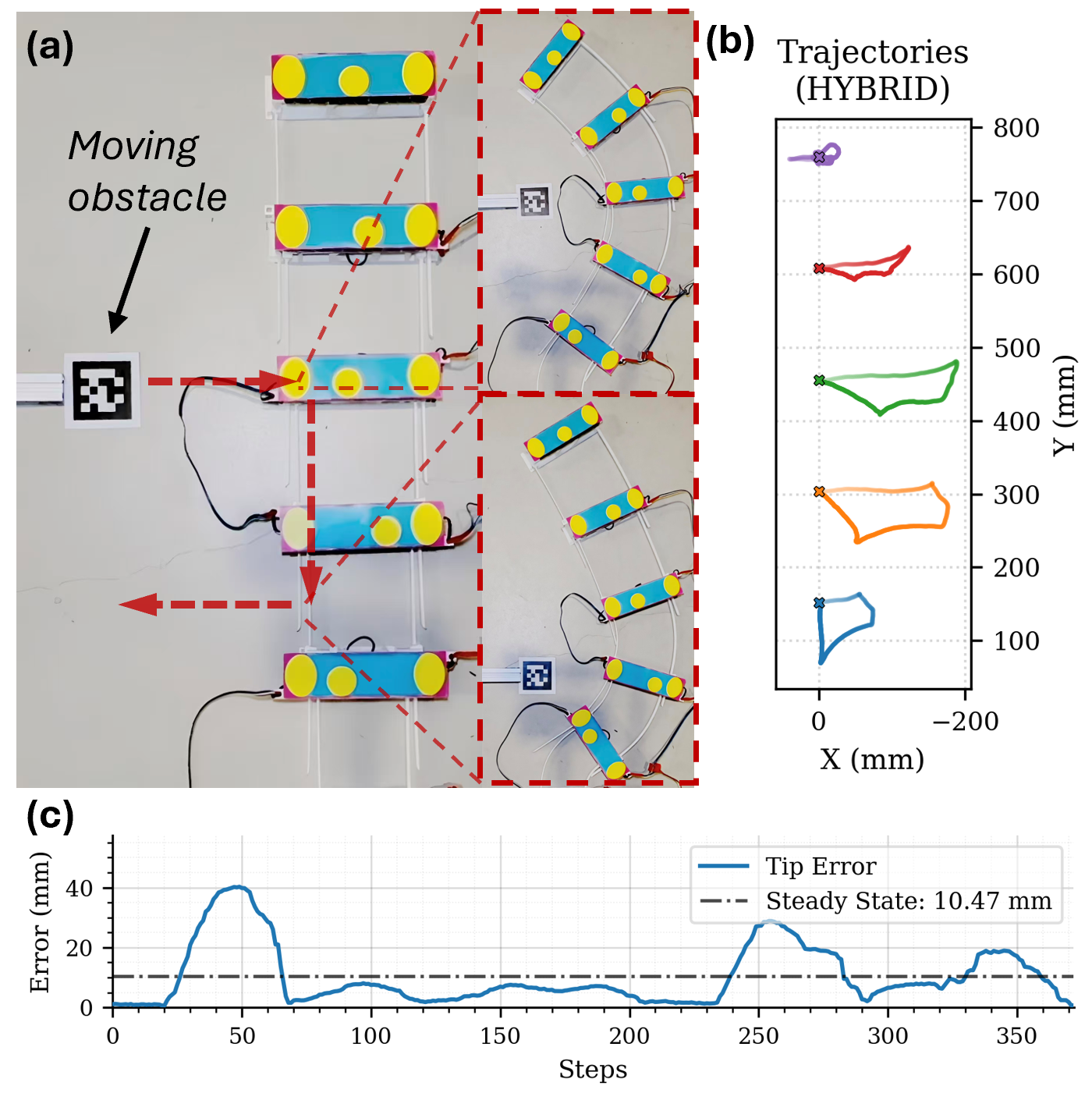} 
    \caption{\textbf{Real-time Null-Space Obstacle Avoidance.} (a) Sequential snapshots of the physical demonstration. As a moving obstacle intrudes from the left into the robot's original workspace, the mechanism dynamically reconfigures its backbone to avoid collision while strictly maintaining the target end-effector pose. (b) The spatial trajectories of all five nodes during the avoidance task. The color gradient (transitioning from light to dark) represents the temporal progression of the movement. (c) The tracking error of the tip position over the control steps. 
    }
    \label{fig:obstacle_avoidance}
\end{figure}

In this experimental setup, the robot uniformly initializes in a highly extended, straight posture with all ten actuation components set to $q = 100$. A vision-based tracking system continuously localizes a dynamic obstacle (represented by an AprilTag \cite{olson2011apriltag}) intruding into the original workspace. To generate real-time avoidance trajectories, an Artificial Potential Field (APF) \cite{khatib1986real} coupled with an envelope boundary method is employed. This planner optimizes a target shape that strictly enforces a fixed tip-pose hard constraint while balancing two underlying objectives: a repulsive force centered on the obstacle to ensure spatial clearance, and a restoring force that continuously encourages the mechanism to return to its nominal straight configuration ($q=100$). 

As illustrated in Fig. \ref{fig:obstacle_avoidance}(a) and (b), the HYBRID framework successfully drives the robot to converge toward these dynamically optimized trajectories, flexibly weaving its backbone around the moving obstacle. During active avoidance, the APF generates target shapes characterized by continuous and natural curvature transitions. However, the task design also inherently introduces a significant elastostatic challenge: as the obstacle moves or recedes, the restoring force continuously drives the mechanism back toward its nominal straight posture. Transitioning back into this near-neutral, low-curvature configuration inevitably causes a loss of sustained structural tension, introducing severe parasitic translations and non-uniform curvatures into the physical system.

Despite these complex elastostatic conditions throughout the active avoidance and recovery phases, the HYBRID controller demonstrates remarkable robustness. The physical structure is undeniably affected by these unmodeled disturbances; however, the SpatioCoupledNet successfully overcomes them by dynamically shifting control authority to the learned spatial predictions when analytical priors degrade. Quantitatively, as shown in Fig. \ref{fig:obstacle_avoidance}(c), the controller ensures high-fidelity null-space manipulation. It bounds the transient tip-positioning error to a peak of 40.02~mm during aggressive obstacle intrusions and maintains a highly precise overall average tracking error of 10.47~mm. Ultimately, the tip error converges to an exceptionally low value at the conclusion of the experiment. This validates the framework's capability to stably harness kinematic redundancy for dynamic collision avoidance while actively compensating for complex boundary-state non-linearities.

\section{CONCLUSION AND FUTURE WORK}

In this paper, we presented a kinematics-informed hybrid closed-loop control framework to overcome the unmodeled effects and inter-segment spatial coupling inherent in hyper-redundant continuum robots. At its core is SpatioCoupledNet, a hierarchical neural architecture that leverages a bidirectional recurrent structure to explicitly model physical force propagation and mechanical hysteresis along the robot backbone. A dimension-wise confidence-gating mechanism enables the controller to dynamically interpolate between an analytical piecewise-constant-curvature prior and a data-driven neural Jacobian, exploiting prior structure where it is reliable while relying on learned corrections where the model is deficient. Together, these components provide a principled path toward accurate, robust whole-body shape control in highly coupled hyper-redundant system.

Experimental results on a custom five-segment planar robot demonstrate that the proposed framework significantly outperforms both purely analytical and purely data-driven baselines. By anchoring neural predictions with physical priors, the hybrid controller achieved lower steady-state errors, accelerated convergence, and moderated actuation costs, particularly in highly deformed boundary configurations. Specifically, under extreme structural deformation, the hybrid controller achieved a steady-state mean error of 6.55~mm and converged in 222 steps, visibly surpassing both the analytical model (26.76~mm, 311 steps) and the purely neural baseline (7.12~mm, 230 steps). Furthermore, the framework's robustness was validated in a dynamic null-space obstacle avoidance task, successfully maintaining a precise mean tip-positioning error of 10.47~mm despite severe, dynamically shifting elastostatic disturbances.

Future work will focus on three primary directions. First, we plan to scale the proposed framework to more complex robotic platforms with a significantly higher number of segments, which will amplify the non-linear inter-segment coupling and further test the scalability of our approach. Second, we aim to optimize the underlying neural architecture by exploring lightweight networks or advanced spatial-modeling paradigms to enhance computational efficiency and feature extraction capabilities. Third, we will improve the system's environmental adaptability by integrating online learning strategies, enabling the controller to autonomously adjust to varying payloads and unpredictable contacts without overfitting to specific laboratory conditions.

\bibliographystyle{IEEEtran}
\bibliography{bib/iros2026}

\end{document}